\documentclass{llncs}
\usepackage{makeidx}  
\usepackage{graphicx}
\usepackage{subfigure}
\usepackage{amssymb,amsmath,bm}
\usepackage{mathrsfs}
\usepackage{booktabs}
\usepackage{multirow}
\begin{document}
	\frontmatter          
	\title{Pyramid Network with Online Hard Example Mining for Accurate Left Atrium Segmentation}
		
	\author{Cheng Bian$^{1}$ \and Xin Yang$^{2,3*}$ \and Jianqiang Ma$^{1}$ \and Shen Zheng$^{1}$ \and Yu-An Liu$^{1}$ \and \\Reza Nezafat$^{3}$ \and
		 Pheng-Ann Heng$^{2}$ \and Yefeng Zheng$^{1}$}
	\institute{Tencent YouTu Lab\and
		Dept. of Computer Science and Engineering, The Chinese University of Hong Kong\\
		\textit{xinyang@cse.cuhk.edu.hk}\and
		Department of Medicine (Cardiovascular Division), Beth Israel Deaconess Medical Center and Harvard Medical School}
	
	\maketitle
	\let\thefootnote\relax\footnotetext{*The work was partly finished when Xin Yang was an internship in the Department of Medicine (Cardiovascular Division), Beth Israel Deaconess Medical Center and Harvard Medical School.}
	\addtocounter{footnote}{-1}\let\thefootnote\svthefootnote
	
	\begin{abstract}
		Accurately segmenting left atrium in MR volume can benefit the ablation procedure of atrial fibrillation. Traditional automated solutions often fail in relieving experts from the labor-intensive manual labeling. In this paper, we propose a deep neural network based solution for automated left atrium segmentation in gadolinium-enhanced MR volumes with promising performance. We firstly argue that, for this volumetric segmentation task, networks in 2D fashion can present great superiorities in time efficiency and segmentation accuracy than networks with 3D fashion. Considering the highly varying shape of atrium and the branchy structure of associated pulmonary veins, we propose to adopt a pyramid module to collect semantic cues in feature maps from multiple scales for fine-grained segmentation. Also, to promote our network in classifying the hard examples, we propose an Online Hard Negative Example Mining strategy to identify voxels in slices with low classification certainties and penalize the wrong predictions on them. Finally, we devise a competitive training scheme to further boost the generalization ability of networks. Extensively verified on 20 testing volumes, our proposed framework achieves an average Dice of $92.83\%$ in segmenting the left atria and pulmonary veins.
	\end{abstract}
	
	\section{Introduction}
	Atrial fibrillation is the most common cardiac electrical disorder, which happens around left atrium and becomes a major cause of stroke \cite{peng2016review}. Ablation therapy guided by MR imaging is a popular treatment solution for atrial fibrillation. Segmenting left atrium in MR scan can benefit the preoperative assessment, determine the catheter size in pulmonary veins, optimize the therapy plan and reduce fluoroscopy time. However, confined by low efficiency and reproducibility, manually segmenting the left atrium tends to be intractable \cite{karim2013evaluation}. \par
	
	\begin{figure}[h]
		\centering
		\includegraphics[width=0.9\linewidth]{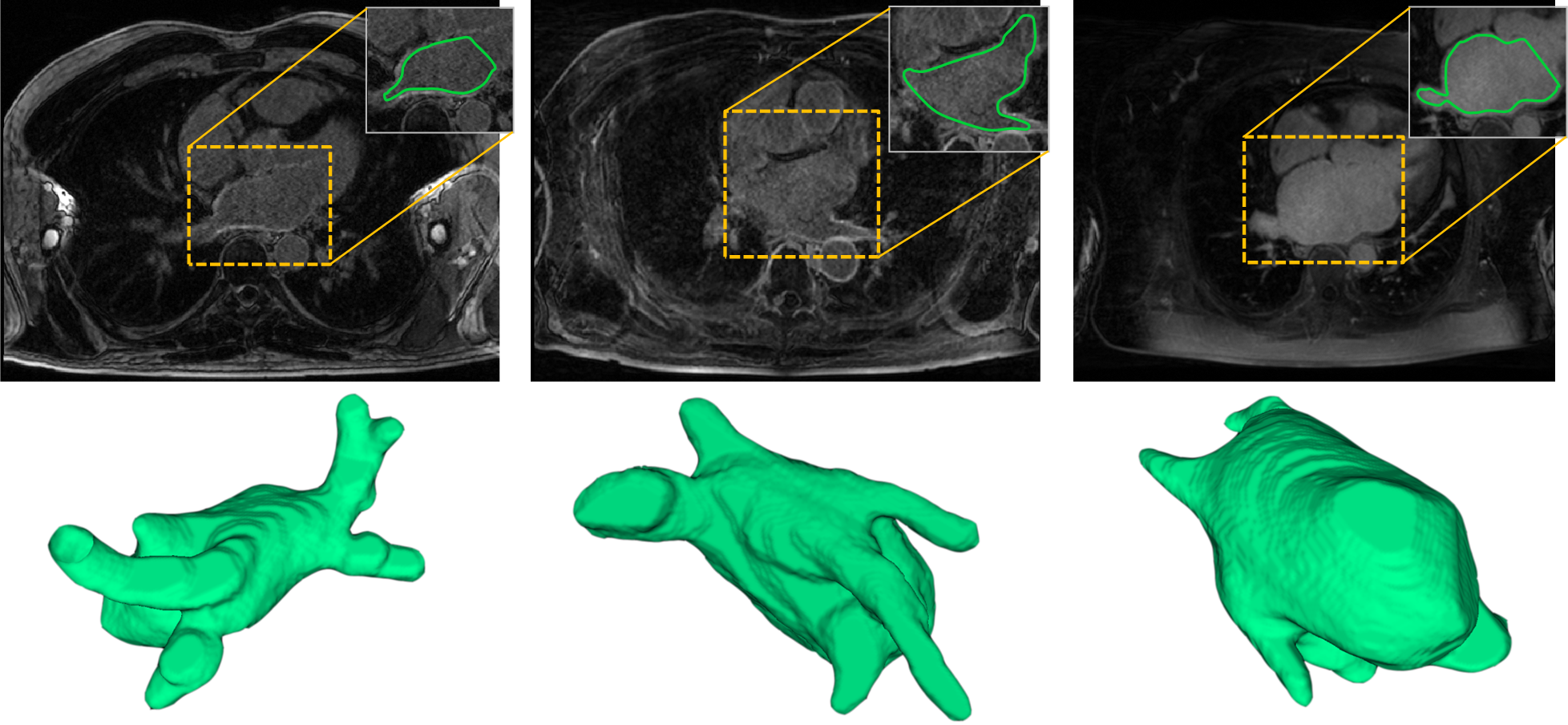}
		\caption{Challenges in segmenting the left atrium and associated pulmonary veins. Green curve and surface rendering denote segmentation ground truth.}
		\label{fig:challenge}
	\end{figure}
	
	As illustrated in Fig. \ref{fig:challenge}, except the poor image quality, automatically segmenting left atrium with pulmonary veins in MR volume is very challenging. Firstly, left atrium only occupies a small ratio when compared with the background, which makes it difficult for algorithm to localize and recognize the boundary details. Secondly, due to the similar intensity with surrounding chambers and the thin myocardial wall with respect to the limited MR resolution, boundary ambiguity and deficiency often occur around the atrium and pulmonary veins. Thirdly, the shape and size of left atrium vary greatly across different subjects and timepoints. Also, the topological arrangement of pulmonary veins, especially the number of veins, presents weak pattern in population \cite{tobon2015benchmark}. \par
	
	Automated left atrium segmentation retains intensive research interest. Deformable model \cite{zheng2014multi}, atlas based label fusion \cite{tao2016fully} and non-rigid registration \cite{zhuang2010registration} are typical methods for atrium included cardiovascular structure segmentation. However, these methods depend heavily on the hand-crafted statistical modeling and lack generalization ability against unseen cases, such as the left atrium with rare number of pulmonary veins. In the deep learning era, this segmentation task is embracing new opportunities. Stacked sparse auto-encoders \cite{yang2017fully}, Fully Convolutional Network \cite{mortazi2017cardiacnet} and Convolutional LSTM \cite{chen2018multiview} were deployed in a 2D fashion for left atrium segmentation. On the other hand, networks in 3D fashion have also been explored \cite{yang2017hybrid}. The superiority between 2D and 3D fashion is arguable and case-by-case \cite{bernard2018deep}. Although networks in both 2D and 3D achieved remarkable performance, preserving segmentation details in varying scales and tackling classification ambiguity around boundary are overlooked in previous work. \par	
	
	In this paper, we propose a deep network based solution for automated left atrium segmentation in gadolinium-enhanced MR (GE-MR) volumes with promising performance. With investigations on our task, we argue that, networks with 2D architectures can easily outperform their heavy 3D versions with respect to time efficiency and segmentation accuracy. Left atrium body possesses a relative compact structure, while the associated pulmonary veins present to be branchy in varying scales. As this regard, we propose to further tailor our 2D network with pyramid pooling modules to enlarge receptive field ranges and perceive semantic cues from multiple scales. This pyramid module greatly promotes the segmentation in fine scales. To avoid the learning of non-informative negative samples and enhance the learning of hard negative examples around boundary, we propose an Online Hard Negative Example Mining strategy to identify voxels in slices with low classification certainties and penalize the wrong predictions on them. Finally, we devise a competitive training scheme in which several competitors need to compete with each other. With the scheme, the generalization ability of all competitors can be further improved. Extensively verified on 20 testing volumes, our proposed framework achieves an average Dice of $92.83\%$ in segmenting the left atria and pulmonary veins.
	\begin{figure}[h]
		\centering
		\includegraphics[width=1.0\linewidth]{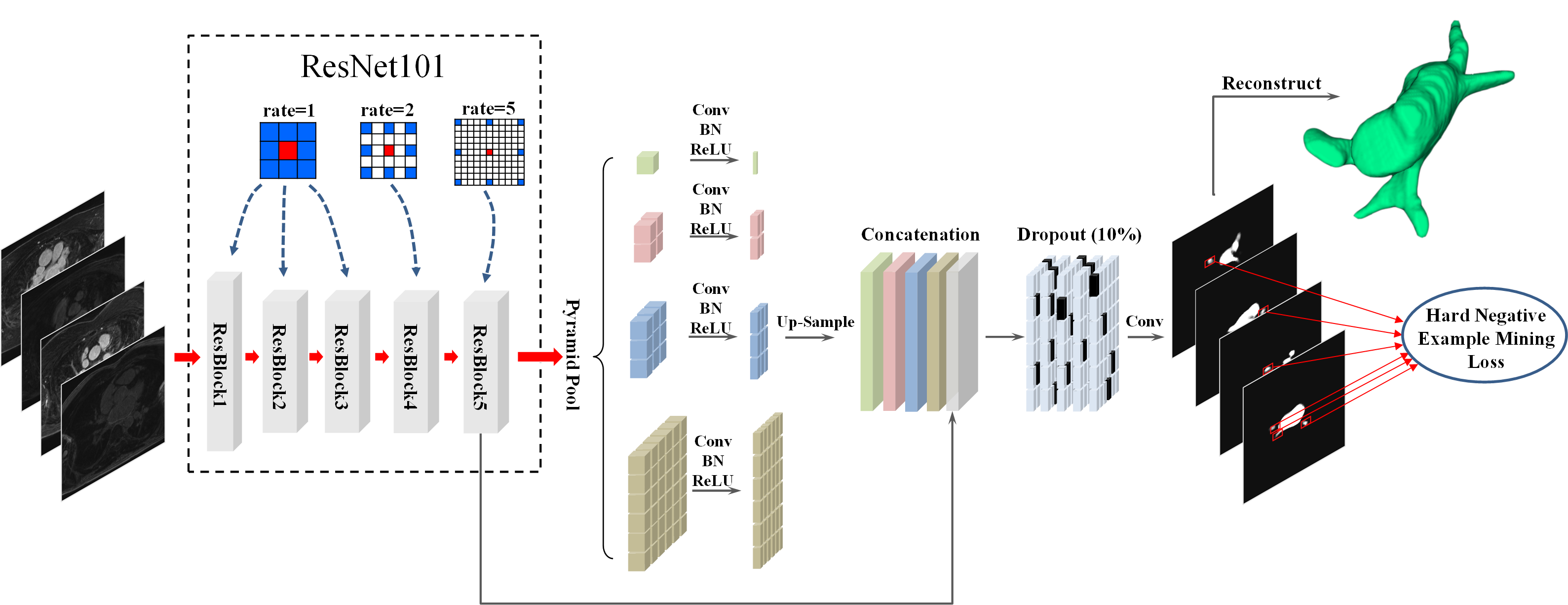}
		\caption{Schematic view of our proposed framework.}
		\label{fig:framework}
		\vspace{-0.5mm}
	\end{figure}
	
	\section{Methodology}
	Fig. \ref{fig:framework} is the schematic view of our proposed framework. Without extra localization modules, our system takes the whole 2D slice as input, hierarchical features are firstly learned and extracted by consecutive residual blocks. Residual blocks are supervised by an auxiliary branch. Then, the adopted pyramid pooling module distills semantic cues from multiple scales. These semantic information are then merged to infer the final predictions. At the end of the network, we introduce the Online Hard Negative Example Mining loss to guide the network selectively search samples in order to learn more efficiently and effectively. \par
	
	\subsection{Choice between 2D and 3D Architecture}
	For volumetric segmentation tasks, 2D and 3D architectures are two main streams and have their different strengths \cite{bernard2018deep,dou20173d}. We investigate several state-of-the-art 2D and 3D architectures for the left atrium segmentation and choose the 2D fashion for following reasons. Firstly, operating in 2D endows our method with super time efficiency, which is about 4 times faster than the 3D version. Secondly, under limited GPU memory, 2D network can get more spaces to expand and deepen its layout which is crucial for the learning capacity of network. Thirdly, 2D network can get large enough and diverse receptive fields to aggregate the in-plane, multiscale context cues for more detailed segmentation. \par
	
	\subsection{Pyramid Network Design}
	Directly applying vanilla 2D networks to segment the left atrium performs poor. The atrium is very small when compared with the background, strong context information, such as the lung and other heart chambers, are needed to localize it. This requires the network have large receptive field and ability to extract proper feature hierarchy. Also, the deformation of atrium and branchy structure of pulmonary veins demand the network can recognize boundaries from macro to nano. \par
	
	Therefore, as shown in Fig. \ref{fig:framework}, we propose to adopt the PSPNet network \cite{zhao2017pyramid} as a workhorse to fit our task. Specifically, for the forehead part, we deploy the ResNet101 pre-trained on ImageNet to extract features. To enhance the network with large receptive fields under parameter limitations, we also utilize the dilated convolution in the ResBlock. For the pyramid pooling module, based on the intermediate feature maps generated by ResBlocks, we use 4 pooling layers to aggregate context information from global to fine scales. The bin size of these four levels are set as 1$\times$1, 2$\times$2, 3$\times$3 and 6$\times$6, respectively. We choose the Atrous Spatial Pyramid Pooling as the pooling kernels. The feature maps produced by the pyramid pooling module are then upsampled and concatenated with the original feature maps from ResBlock. A last convolution layer is applied on the concatenated feature maps to output the final predictions. Noted that, we apply the dropout on this last convolution layer as an ensemble strategy to improve the generalization ability of our network. Dropout ratio is set as 0.1. \par
	
	Same with \cite{zhao2017pyramid}, we inject an auxiliary supervision branch at the last ResBlock to tackle the potential training difficulties faced by deep neural networks. As a way to tackle the class imbalance, we use the weighted cross entropy as the basic form of our objective function. \par
	
	\subsection{Online Hard Negative Example Mining}\label{section:OHEM}
	For network training, the definition of objective function directly defines the latent mapping that network needs to fit. The quality of samples counted to minimize the objective function can determine how close the network can achieve its goal. In this work, we find that there is an obvious imbalance between the amount of foreground and background samples within slices, which can bias the loss function. Also, most background samples are easy to be classified and only some voxels around the atrium boundary get high probabilities to be false alarms. Therefore, we propose an Online Hard Negative Example Mining (OHNEM) strategy to tune the objective function and selectively search informative hard negative samples to significantly improve the training efficacy. \par
	
	Motivated by \cite{shrivastava2016training}, our OHNEM strategy roots in three facts, 1) most negative samples are becoming non-informative as the training progresses and should be excluded from the loss to alleviate the foreground-background imbalance, 2) hard negative examples are those getting high prediction scores to be foreground, 3) we preserve all positive samples to learn the knowledge of foreground thoroughly. In this regard, our OHNEM consists of three major steps in each training iteration. First, determining the number of hard negative examples to be selected. We adaptively define this number as $C_{hn}=max(2*C_p, min(5, C_n/4))$, where $C_p$ and $C_n$ are the number of foreground and background voxels obtained from the label slice, respectively. Then, all the background samples are ranked in a descending order according to their prediction scores. Finally, our loss function only counts the top $C_{hn}$ background candidates and discards the rest. Our loss function equally considers the losses from all the foreground samples and the selected background samples. Effectiveness of the proposed OHNEM is verified with ablation study in Section \ref{section:experiment}. \par

	\begin{figure}[h]
		\centering
		\includegraphics[width=0.85\linewidth]{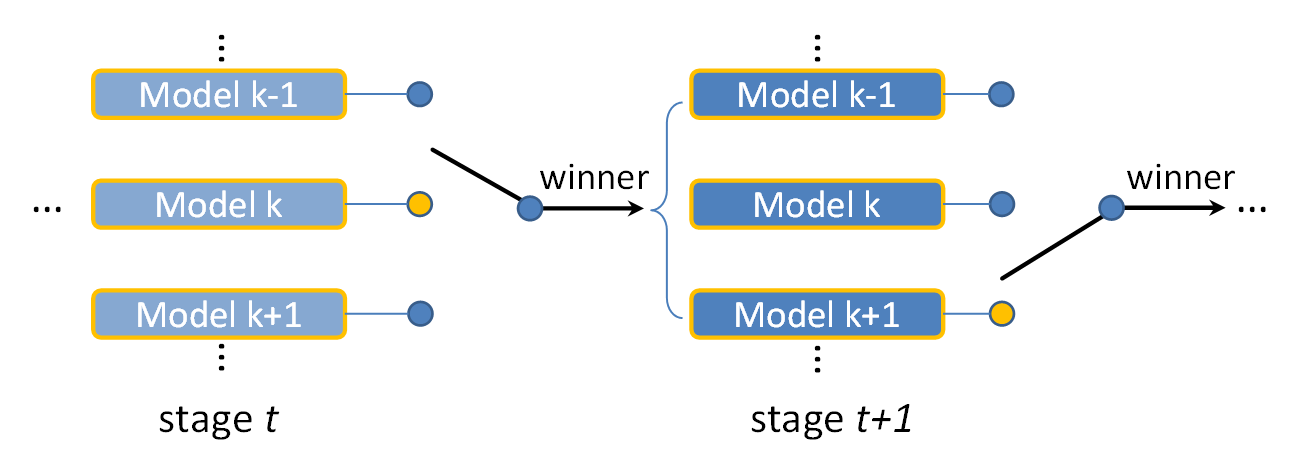}
		\caption{The proposed competitive training scheme with multiple model candidates. Winner model in each stage is denoted with yellow dot.}
		\label{fig:competitive}
		\vspace{-0.5mm}
	\end{figure}
	
	\subsection{Competitive Training Scheme} 
	Because of the random conditions in training, networks with the same training setting can have different generalization abilities. An ensemble of multiple networks often surpasses the single network. In this paper, we propose a new training scheme to generate an ensemble of multiple networks in a competitive way, denoted as CTS. As shown in Fig. \ref{fig:competitive}, our CTS training scheme is intuitive in design. The scheme simultaneously trains several competitors with the same initial configurations. Only the winner model with the lowest validation loss can be recorded at the validation point. The winner model in stage $t$ can get the chance to broadcast its parameters to initialize all the competitors in next competitive stage. All the competitors are then be fine-tuned in stage $t+1$ until the generation of a new winner at next validation point. \par
	
	With the CTS training scheme, all competitors are facing with the more and more intensive competition. All candidates are enhanced with the local optimal model after each validation stage. They can only struggle to mine more informative examples for effective gradient update, which in turn improves the performances of all competitors. Also, different from the direct and heavy ensemble of multiple networks for testing, our network is light-weight for testing since only the winner model is kept at the end. \par
	
	\section{Experimental Results} \label{section:experiment}
	\subsubsection{Experiment materials:} We evaluated our method on the Atrial Segmentation Challenge dataset in STACOM 2018. There are two kinds of image size as 576$\times$576$\times$88 and 640$\times$640$\times$88 with the unified spacing 1.0$\times$1.0$\times$1.0mm. We split the whole dataset (100 volumes with ground truth annotation) into 70/10/20 for training/validation/testing. Training dataset is further augmented with random flipping. Every volume is split into slices along the last dimension. Slices are normalized as zero mean and unit variance before training and testing. \par
	
	\subsubsection{Implementation details:} We implement all the compared methods with the \textit{PyTorch} using NVIDIA GeForce GTX TITAN X GPUs. We update the weights of network with an Adam optimizer (batch size=2, learning rate=0.001). All the models are trained for 180 epochs with a validation every 30 epochs. Only the model with the lowest validation loss is saved. For the competitive training, the winner is generated every 30 epochs. For the post-processing in testing, small isolated connected components are removed in the final labeling result. We implemented several state-of-the-art 3D and 2D networks varying with different layouts and kernel designs for extensive comparison. We also conducted ablation study on our proposed modules.
	\begin{table}[!htb] \caption {Quantitative evaluation of our proposed framework} \label{table:net_metric}
		\centering
		\begin{tabular}{c|c|c|c|c|c}
			\toprule[2pt]
			\multirow{2}{*}{\bf{Method}} & \multicolumn{5}{c}{\bf{Metrics}}\\
			\cline{2-6}
								&Dice[\%] 	&Conform[\%]  			&Jaccard[\%]	&Adb[mm] 	&Hdb[mm] 	 \\
			\hline
			PSPNet \cite{zhao2017pyramid}				&87.904		&72.218		&78.532			&2.159			&23.319	\\
			PSPNetD				&92.053		&82.536		&85.387			&1.564			&21.154	\\
			Unet-2D				&89.638		&76.485		&81.404			&2.185			&22.711	\\
			Unet-3D				&87.020		&68.033		&77.785			&2.082			&25.243	\\
			DeepLabV3 \cite{chen2017rethinking}			&88.456		&72.449		&79.846			&2.427			&23.099	\\
			SegNet \cite{badrinarayanan2017segnet}			&90.762		&78.912		&83.422			&2.050			&23.362	\\
			GCN \cite{peng2017large}					&91.812		&82.049		&84.926			&1.597			&18.940	\\
			\hline
			PSPNetD+OHNEM		&92.688		&84.113		&86.434			&\textbf{1.405}			&\textbf{17.224}	\\
			PSPNetD+OHNEM+CTS3	&\textbf{92.834}		&\textbf{84.445}		&\textbf{86.694}		&1.496			&17.891	\\
			\hline
			\toprule[2pt]
		\end{tabular}
		\vspace{-0.5cm}
	\end{table}
	
	\subsubsection{Quantitative and Qualitative Analysis:}
	We use 5 metrics to evaluate all the compared solutions, including Dice, Conform, Jaccard, Average Distance of Boundaries (Adb) and Hausdorff Distance of Boundaries (Hdb). We first compare our core network \textit{PSPNet} with other architectures. \textit{PSPNet} with dilated convolution and dropout is denoted as \textit{PSPNetD}. We can observe that, sharing the same spirit in modifying kernels to perceive broad context, both \textit{PSPNetD} and \textit{GCN} get better results than other architectures which use vanilla convolutions. \textit{PSPNetD} performs even better than \textit{GCN}, which may because of the explicit exploration on multiscale semantic cues. \textit{Unet-3D} loses power in our task and present poor results than all 2D networks. \par

	\begin{figure}[h]
		\centering
		\includegraphics[width=1.0\linewidth]{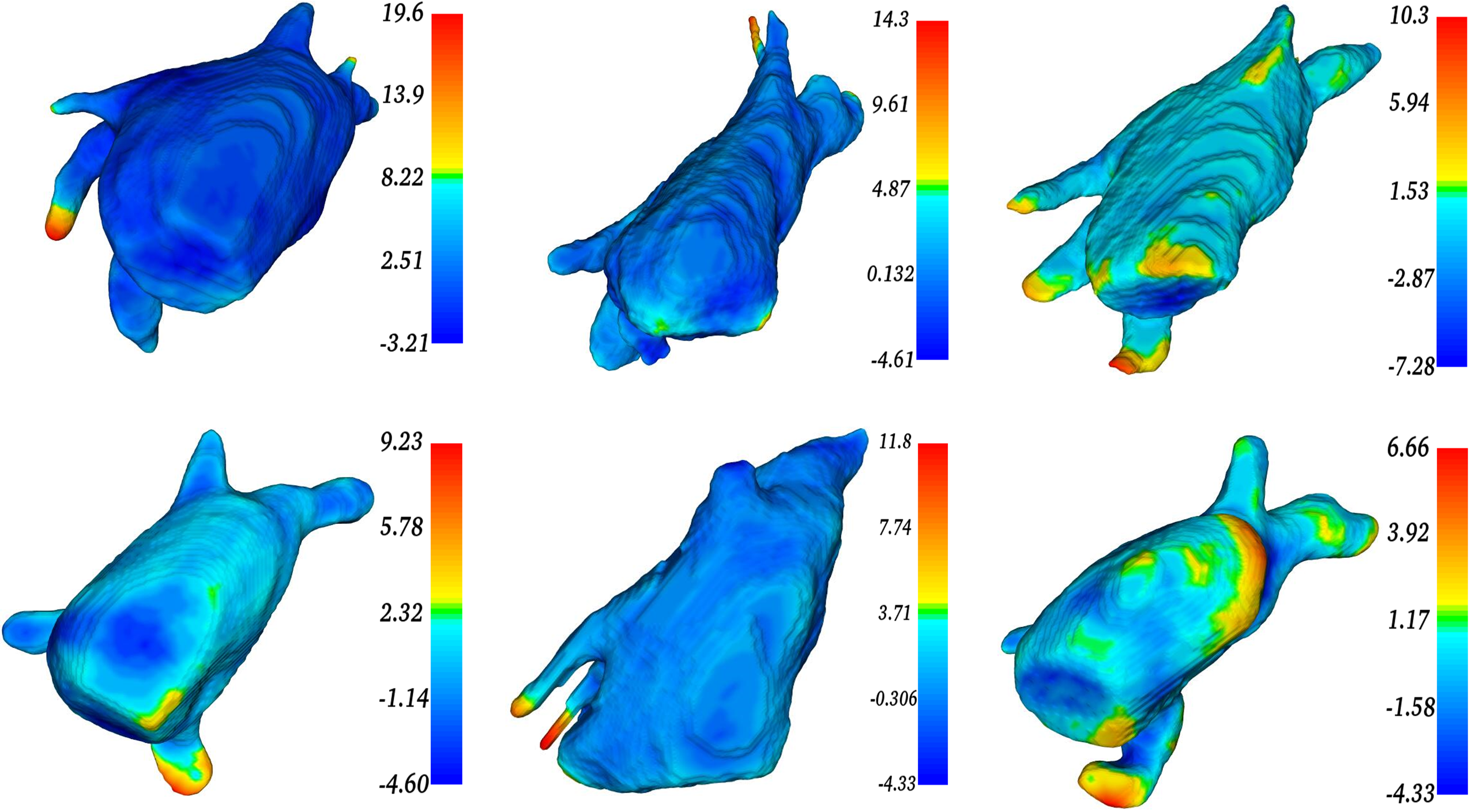}
		\caption{Visualization of our segmentation results with the mapping of Hausdorff distance. Color bar is annotated with mean in the center, min and max on the ends.}
		\label{fig:seg_results}
	\end{figure}
	
	Based on \textit{PSPNetD}, we conduct ablation study on OHNEM and CTS. There occurs a significant improvement (about 0.6$\%$ in Dice) when OHNEM module is involved to guide the learning of PSPNetD (denoted as \textit{PSPNetD+OHNEM}). This proves the importance of designing proper objective function and selecting effective samples for training. For the competitive training scheme CTS, we can see that, with 3 candidates for competition, \textit{PSPNetD+OHNEM+CTS3} brings another 0.16$\%$ improvement in Dice over \textit{PSPNetD+OHNEM}. The learning of hard negative samples is already enhanced in \textit{PSPNetD+OHNEM}, while CTS further pushes candidate models for more effective learning. \par
	
	Finally, we visualize the segmentation results of \textit{PSPNetD+OHNEM+CTS3} in Fig. \ref{fig:seg_results} with Hausdorff distance [mm] between segmentation and ground truth. Our method presents robustness against the shape and size variations in left atrium and pulmonary veins. Most points on segmentation surface around the atrium body present low Hausdorff distances (cold color) to the ground truth. Large distance values (hot color) often occur around the end of pulmonary veins which are hard to define since the length of of pulmonary vein varies greatly. \par
	
	\section{Conclusions}
	In this paper, we present a fully automatic framework for the segmentation of left atrium and pulmonary veins in GE-MR volume. We start with network architecture in 2D, and adopt kernels for large receptive field. The involvement of pyramid pooling improves our network in combing global and local cues to recognize detailed boundaries. The proposed Online Hard Negative Example Mining significantly boosts the training efficacy. Competitive training scheme pushes the network for further improvement. Verified on 20 testing cases, our method presents promising performance in segmenting left atrium and pulmonary veins across different subjects and imaging conditions. \par
	
	\section{Acknowledgments:}
	The work in this paper was supported by a grant from the Research Grants Council of the Hong Kong Special Administrative Region (Project no. GRF 14225616). \par
	
	%
	\bibliographystyle{splncs}
	\bibliography{stacom_refs}	
	
\end{document}